\documentclass[conference]{IEEEtran}
\IEEEoverridecommandlockouts
\usepackage{cite}
\usepackage{amsmath,amssymb,amsfonts}
\usepackage{algorithmic}
\usepackage{graphicx}
\usepackage{textcomp}
\usepackage{xcolor}
\def\BibTeX{{\rm B\kern-.05em{\sc i\kern-.025em b}\kern-.08em
    T\kern-.1667em\lower.7ex\hbox{E}\kern-.125emX}}

\usepackage{color}

\usepackage[utf8]{inputenc} 
\usepackage[T1]{fontenc}    
\usepackage{hyperref}       
\usepackage{url}            
\usepackage{booktabs}       
\usepackage{amsfonts}       
\usepackage{nicefrac}       
\usepackage{microtype}      
\usepackage{xcolor}         

\usepackage{wrapfig}

\usepackage{hyperref}
\usepackage[nameinlink]{cleveref}

\hypersetup{
  linkcolor=black,       
  citecolor=black,       
  filecolor=black,       
  urlcolor=black,        
  colorlinks=true,     
}

\begin{document}

\title{Transformer-based Graph Neural Networks for Battery Range Prediction in AIoT \\ Battery-Swap Services

\thanks{\IEEEauthorrefmark{5}Equal contribution.}
\thanks{\IEEEauthorrefmark{1}Corresponding author.}
}

\author{\IEEEauthorblockN{Zhao Li\IEEEauthorrefmark{8}\IEEEauthorrefmark{6}\IEEEauthorrefmark{5},
Yang Liu\IEEEauthorrefmark{2}\IEEEauthorrefmark{5},
Chuan Zhou\IEEEauthorrefmark{2}\IEEEauthorrefmark{9}\IEEEauthorrefmark{1}, 
Xuanwu Liu\IEEEauthorrefmark{6}
Xuming Pan\IEEEauthorrefmark{6}
Buqing Cao\IEEEauthorrefmark{3}\IEEEauthorrefmark{1} and 
Xindong Wu\IEEEauthorrefmark{4}
}
\IEEEauthorblockA{\IEEEauthorrefmark{8}Zhejiang Lab, Hangzhou, China}
\IEEEauthorblockA{\IEEEauthorrefmark{6}Hangzhou Yugu Technology Co., Ltd, Hangzhou, China}
\IEEEauthorblockA{\IEEEauthorrefmark{2}Academy of Mathematics and Systems Science, Chinese Academy of Sciences, Beijing, China}
\IEEEauthorblockA{\IEEEauthorrefmark{9}School of Cyber Security, University of Chinese Academy of Sciences, Beijing, China}
\IEEEauthorblockA{\IEEEauthorrefmark{3}Hunan University of Science and Technology, Xiangtan, China}
\IEEEauthorblockA{\IEEEauthorrefmark{4}Hefei University of Technology, Hefei, China\\
Email: lzjoey@gmail.com, \{liuyang2020, zhouchuan\}@amss.ac.cn, \{liuxuanwu, panxm\}@yugu.net.cn,\\ buqingcao@gmail.com, xwu@hfut.edu.cn}
}

\maketitle

\begin{abstract}
The concept of the sharing economy has gained broad recognition, and within this context, Sharing E-Bike Battery (SEB) have emerged as a focal point of societal interest. Despite the popularity, a notable discrepancy remains between user expectations regarding the remaining battery range of SEBs and the reality, leading to a pronounced inclination among users to find an available SEB during emergency situations.  In response to this challenge, the integration of Artificial Intelligence of Things (AIoT) and battery-swap services has surfaced as a viable solution. In this paper, we propose a novel structural Transformer-based model, referred to as the SEB-Transformer, designed specifically for predicting the battery range of SEBs. The scenario is conceptualized as a dynamic heterogeneous graph that encapsulates the interactions between users and bicycles, providing a comprehensive framework for analysis. Furthermore, we incorporate the graph structure into the SEB-Transformer to facilitate the estimation of the remaining e-bike battery range, in conjunction with mean structural similarity, enhancing the prediction accuracy. By employing the predictions made by our model, we are able to dynamically adjust the optimal cycling routes for users in real-time, while also considering the strategic locations of charging stations, thereby optimizing the user experience. Empirically our results on real-world datasets demonstrate the superiority of our model against nine competitive baselines. These innovations, powered by AIoT, not only bridge the gap between user expectations and the physical limitations of battery range but also significantly improve the operational efficiency and sustainability of SEB services. Through these advancements, the shared electric bicycle ecosystem is evolving, making strides towards a more reliable, user-friendly, and sustainable mode of transportation.
\end{abstract}


\section{Introduction}


The emergence of the sharing economy has become ubiquitous across diverse facets of modern society \cite{cheng2016sharing,hossain2020sharing,li2024real}, spotlighting Sharing E-Bike Battery (SEB) \cite{li2021exploring} as a focal point of attention. Furthermore, the sharing battery assumes a central role in numerous intelligent systems, particularly in the context of AIoT with battery-swap services. As an illustration, in e-bike battery-swap services , range prediction functions as the primary propulsion source, delivering the necessary driving force for the entire vehicle’s operation. The performance of SEB plays a pivotal role in determining critical factors, including driving range \cite{lu2020optimal}, fuel efficiency \cite{kiray2021significant}, and safety standards \cite{yu2023evaluation} within the realm of electric vehicles. Additionally, a noticeable disparity persists between user expectations and the actual remaining range of SEBs. This discrepancy is particularly evident when users urgently require access to available SEBs, a situation that carries significant implications for alleviating range-related concerns during their journeys. The provision of accurate range prediction and strategically optimal placement of exchange stations can greatly aid users in their route planning endeavors \cite{bast2016route,alberts2007planning,krishna2023ai}, thereby enhancing operational efficiency and promoting energy conservation initiatives. 

\begin{figure*}[t]
    \centering
    \includegraphics[width=0.8\linewidth]{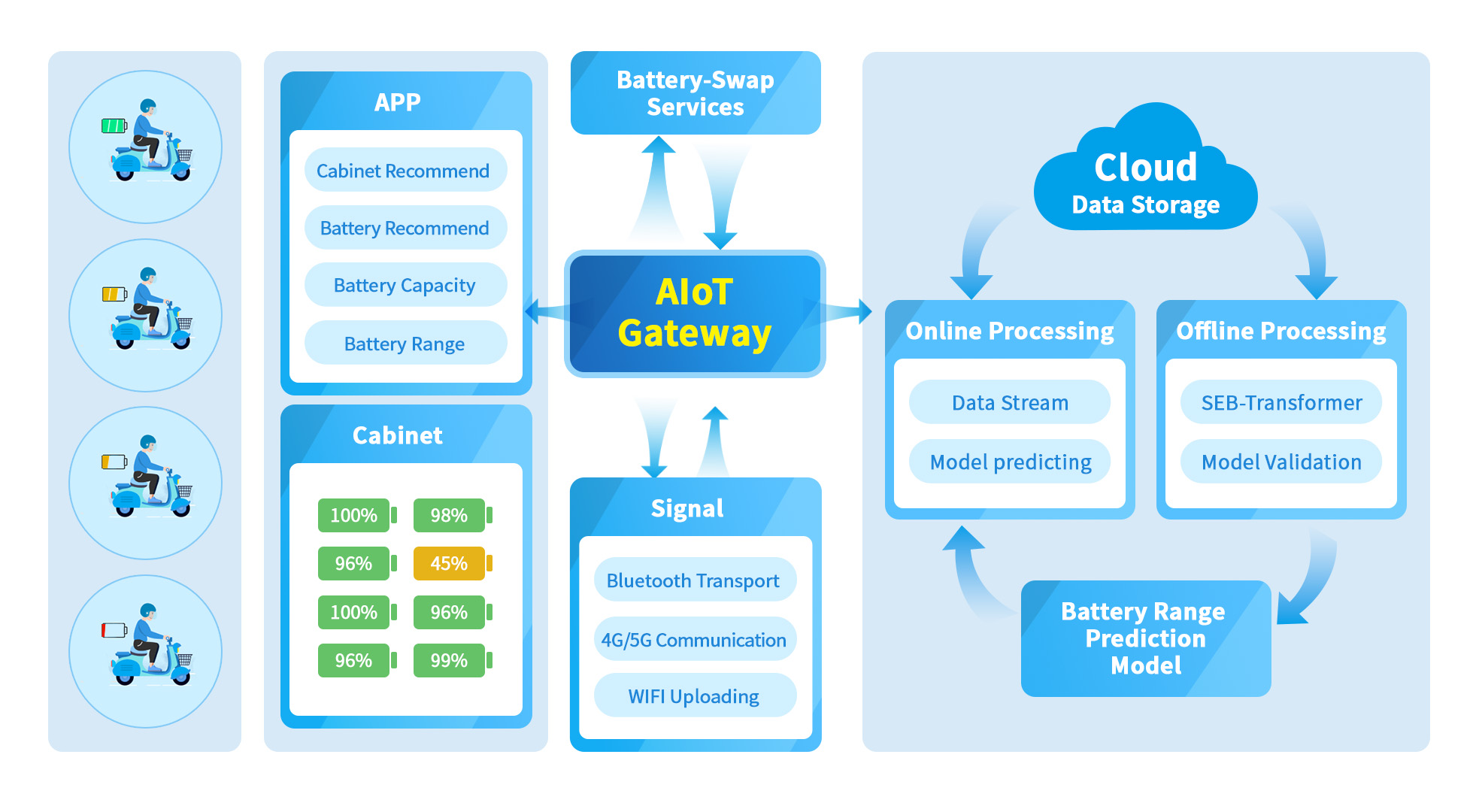}
    \caption{\small{AIoT Battery-Swap Services. The workflow involves riders interacting with AIoT systems through an app, accessing battery-swap services designed for efficiency and convenience. AIoT system integration includes three key aspects: signal and sensor interfacing for real-time data stream, cloud center and transformer-based model for advanced processing and analytics, and finally, services seamlessly connecting with the app to deliver user-centric solutions.}} \label{fig:AIoT} 
\end{figure*}

{\bf Range anxiety}\;of riders need to swap batteries primarily due to insufficient battery range, as e-bikes can only travel a limited distance on a single charge. This necessitates swapping when the battery depletes. Additionally, long charging times contribute to the need for swapping, as waiting for a battery to recharge is often impractical, especially for riders using e-bikes for delivery services or commuting. Battery degradation over time also means batteries hold less charge, requiring more frequent swaps. Operational efficiency, particularly for commercial use, favors swapping over recharging to ensure continuous operation. Convenience plays a significant role, as swapping stations offer a quick solution compared to the lengthy recharging process. In areas with inadequate charging infrastructure, swapping becomes a practical necessity. Lastly, range anxiety—the fear of running out of power without access to charging—prompts riders to prefer battery swapping to ensure their e-bikes are always ready for use. This study focuses primarily on this aspect, with a specific emphasis on the refinement and enhancement of range prediction accuracy.

{\bf AIoT Battery-Swap Services}\;denote a sophisticated system integrating Artificial Intelligence of Things (AIoT) technology with battery-swapping facilities tailored for e-bike, marking the advent of a revolutionary phase in energy management and mobility solutions (see Fig.\ref{fig:AIoT}). The primary objective of this service is to augment the efficiency and convenience of electric e-bike usage. This is achieved through the utilization of smart sensors, connectivity solutions, and advanced AI algorithms, all geared towards optimizing battery management and exchange processes. These processes encompass battery usage monitoring, demand pattern prediction, and the streamlining of the swapping procedure. A key focal point of AIoT battery-swapping services lies in the enhancement of battery range and longevity. Through the utilization of real-time data analytics and predictive algorithms, AIoT systems possess the capability to accurately evaluate battery health, forecast remaining range, and suggest optimal intervals for battery swapping. Such a proactive approach plays a pivotal role in mitigating range anxiety among e-bike riders, guaranteeing timely access to fully charged batteries whenever required. Consequently, this fosters the widespread adoption of batteries, thus contributing to the acceleration of sustainable transportation solutions. Through the integration of AIoT Battery-Swap Services outlined in our paper, we overcome the constraints associated with traditional charging infrastructure. This results in an array of benefits, including heightened convenience \cite{chang2021survey}, scalability \cite{chang2021survey}, and sustainability, laying the groundwork for a more sustainable and efficient transportation network \cite{alaba2024enabling}. This innovative technology not only expedites the adoption of batteries but also lays the groundwork for a smarter, interconnected ecosystem poised to revolutionize the future of transportation and energy consumption, ushering in an era of sustainable and efficient mobility solutions.

Considering the complex characteristics inherent in this dynamic prediction task, our research focus is increasingly oriented towards the utilization of Transformers, which are currently recognized extensively in the academic community. The Transformer architecture, a groundbreaking development in neural network design, has significantly transformed the fields of natural language processing and sequential learning. Its innovative attention mechanism has established new benchmarks for state-of-the-art performance across a multitude of domains. Interest in investigating data-driven approaches is growing rapidly \cite{hasib2021driving, gallardo2014battery, hu2012multiscale, li2018single, li2019data}, particularly those leveraging Transformer-based models, for the prediction of battery range. However, the exclusive reliance on Transformers or alternative models to address the complexities of dynamic graph sequence tasks is inadequate. This approach fails to account for structural information and the critical interactions between the graph elements and sequences. Transforming non-Euclidean interactive networks \cite{asif2021graph} into Euclidean space represents a significant challenge for traditional machine learning methodologies. Through the adoption of a graph structure, Transformers are adept at capturing the intricate relationships and interactions that exist between batteries and users efficiently. Summarily, our research endeavors to address the following pivotal question: \emph{``How can an advanced neural network architecture be developed to more effectively incorporate structural information for the prediction of battery range, through the innovative application of Transformers?"}

In this study, we delve deeply into the pivotal issue of predicting the remaining range in Sharing E-Bike Battery (SEB), with a specific focus on applications within the Artificial Intelligence of Things (AIoT) domain. This concern is of utmost importance for end-users, influencing both usability and service satisfaction. We propose a novel, task-oriented model termed as SEB-Transformer, which employs a structural Transformer-based methodology specifically devised for the accurate prediction of battery range in SEBs. We envision the scenario as a dynamic, heterogeneous graph, meticulously crafted to encapsulate the intricate interplay and connections between users and bicycles. Additionally, we seamlessly integrate this graph structure into the SEB-Transformer framework. This integration facilitates the concurrent estimation of not only the remaining range of e-bike battery but also the average structural similarity across the system, providing a holistic view of the network's dynamics. By leveraging the insights obtained from our model, we dynamically optimize cycling routes for users in real time. This optimization process considers the spatial distribution of charging stations, thereby enhancing user experience by ensuring efficient energy use and reducing the likelihood of battery depletion during trips.

Our contributions are summarized as follows:
\begin{itemize}
    \item {\bf AIoT Battery-Swap Services.}\;We delineate the scenario of Sharing E-Bike Battery (SEB) as a dynamic heterogeneous graph, a conceptual framework that allows for the representation of complex relationships and attributes. Furthermore, we elucidate the introduction of structural information into this framework, thereby leveraging GNN's powerful capability to process and analyze graph-structured data.
    \item {\bf Transformer-based Structual GNNs.}\;We propose the SEB-Transformer model, a novel structural Transformer-based framework designed specifically for the prediction of sharing e-bike battery range. This model leverages the advanced capabilities of Transformer architectures to analyze and predict battery usage patterns, incorporating a structural perspective that enhances its predictive accuracy and reliability for applications in e-bike.
    \item {\bf Real-world Evaluation.}\;In our empirical analysis, the SEB-Transformer model was evaluated using a dataset derived from real-world scenarios, alongside a comparison against nine other competitive baseline models. The results of this evaluation unequivocally illustrate the enhanced performance of our model. Notably, it outperforms the traditional vanilla Transformer model by a significant margin, achieving an improvement of over 36.7\%. This notable enhancement underscores the effectiveness of the SEB-Transformer in addressing the complexities of range prediction in sharing e-bike battery.
\end{itemize}


\noindent{\bf Outline.}\;Section \ref{section:related} provides an overview of related works, encompassing GNNs, Transformer, and their applications in prediction tasks. Section \ref{section:pre} offers an introduction to the fundamentals of Transformers and GNNs. Subsequently, we present our proposed framework and discuss the scenario involving shared e-bike battery. Section \ref{section: application} delves into various applications linked with web services. Finally, Section \ref{section:exp} presents a comprehensive experimental study conducted on a real-world dataset.

\section{Related Works}\label{section:related}

This section provides an overview of related works, including models utilized in SEB-Transformer, such as GNNs and Transformer. Additionally, we will discuss the connection to our work.

\paragraph{GNNs and Transformer} The fusion of graph neural networks (GNNs) \cite{kipf2017semi,zhang2025surveygraphretrievalaugmentedgeneration,chen2024entity} and Transformers marks a pivotal development in machine learning, with GNNs excelling in graph data analysis and Transformers advancing sequence tasks. Prior to Transformers \cite{min2022Transformer,vaswani2017attention,dwivedi2020generalization,ying2021Transformers}, RNNs, including LSTMs \cite{sherstinsky2020fundamentals} and GRUs \cite{vaswani2017attention}, dominated sequence processing but fell short in distributed computing. The advent of the attention mechanism, epitomized by Google's BERT \cite{devlin2018bert}, revolutionized NLP by enhancing focus on relevant data segments during processing. This synergy between GNNs for local structure and Transformers for global dependencies is now under active investigation, aiming to unify their strengths for improved performance across tasks.

\paragraph{Transformer for prediction task} Deep neural networks have increasingly become the foundational framework for prediction tasks \cite{xu2023multi}, paralleled by the substantial development of the Transformer model. Transformer model is optimally utilized in scenarios where it can simultaneously generate predictions for interrelated tasks. Research into employing Transformers for prediction tasks has become a critical focus within the machine learning domain. Originally designed for natural language processing tasks \cite{min2022Transformer,vaswani2017attention}, Transformers have showcased exceptional ability in identifying complex sequential patterns and managing long-range dependencies. Expanding the use of Transformers \cite{xu2023multi,nambiar2020transforming,yang2020html} beyond traditional sequence-based applications to include a variety of domains like time-series forecasting, image classification, and financial prediction has broadened their applicability. This study delves into the use of Transformers in predictive modeling, highlighting their strengths in deciphering complex data relationships.


\paragraph{Transformer for combinatorial task} Beyond the widespread adoption of the simulated annealing technique for a range of combinatorial optimization challenges \cite{wang2023solving,yildiz2022reinforcement,wanggraph}, recent scholarly efforts have explored the integration of RNNs in this sphere. Although RNNs have not uniformly achieved error rates as minimal as those attained through annealing methods, the findings have been promising, showcasing significant improvements in solution speed. This suggests a strategic equilibrium between precision and high efficiency, a compromise that might be particularly appealing to stakeholders like Google Maps users and crew schedulers seeking to refine their scheduling processes. The exploration of Transformer models for combinatorial tasks has become a focal point in contemporary machine learning research. Initially conceived for natural language processing tasks, Transformers have demonstrated exceptional proficiency in identifying complex dependencies and patterns, prompting their adoption in a variety of fields, such as combinatorial optimization challenges. Combinatorial tasks fundamentally involve choosing the best solutions from a set of discrete options, making them prevalent in areas like logistics \cite{ren2023review}, operations research \cite{mazyavkina2021reinforcement}, and network design \cite{chen2013markov}.

\section{Preliminaries}\label{section:pre}

\noindent{\bf Notation.}\;We denote a graph as $\mathcal{G}$ and  its edges and nodes as $\mathcal{E}$ and $\mathcal{V}$ respectively. 
We represent an ordinary graph as a set of edges $\mathcal{E} = \{ (v_i, v_j) \mid v_i, v_j \in \mathcal{V} \}$, where $n$ is the number of observed edges. For each node $v$ and edge $e$, we use their bold version $\mathbf{v}$ and $\mathbf{e}$ to denote their embeddings. We use bold capital letters. e.g., $\mathbf{A}, \mathbf{B}, \mathbf{W}$ to denote matrices and use $\| \cdot \|$ to denote the Euclidean norm of vectors or the Frobenius norm of matrices. Due to space limitations, we summarize this paper's main symbols and notations in Table \ref{table: notation}.

\begin{table}[t]
\small
    \caption{\small{Notation table.}} \label{table: notation}
    \vskip 0.05in
    \centering
         \begin{tabular}{lp{4cm}} 
        \toprule
        Name & Description \\
        \midrule
        $t$ & time step \\
        $HG^{(t)} = (V^{(t)}, E^{(t)})$ & heterogeneous graph at time $t$ \\
        $V^{(t)} = \{v_i^{(t)}\}_{i=1}^N$ & set of nodes at time $t$ \\
        $E^{(t)} = \{e_i^{(t)} = (v_{i,1}^{(t)}, v_{i,2}^{(t)})\}_{i=1}^M$ & set of edges at time $t$ \\
        $\mathcal{T} = \{t_i\}_{i=1}^T$ & set of node's type \\
        $\mathcal{R} = \{r_i\}_{i=1}^R$ & set of edge's type \\ \midrule
        $\mathbf{W}, \mathbf{B}$ & parameters in GNNs\\
        $\mathbf{Q}, \mathbf{K}, \mathbf{V}$ & query/key/value matrix in attention mechanism\\
        \bottomrule
         \end{tabular}
\end{table}

\subsection{Scenario of SEBs}

\begin{figure}[t]
    \centering
    \includegraphics[width=1.0\linewidth]{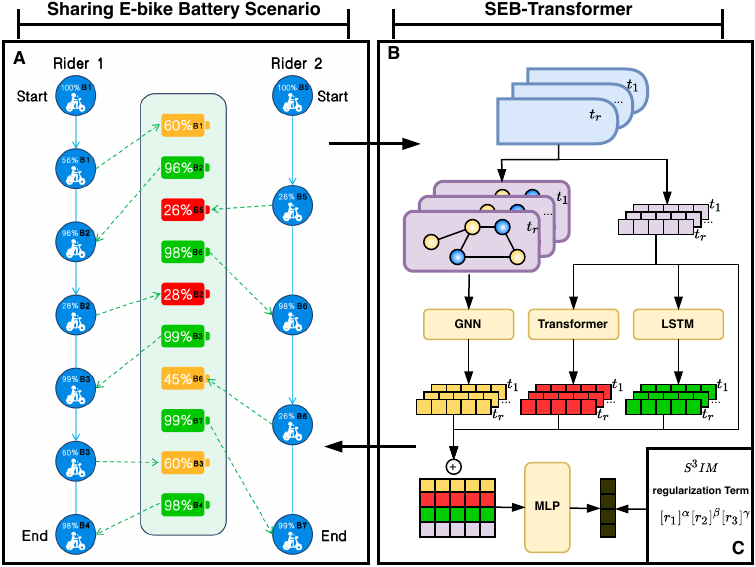}
    \caption{\small{Illustration of SEB scenario and SEB-Transformer.}} \label{fig:SEBs} 
    \vspace{10pt}
\end{figure}

The scenario involving Sharing E-Bike Battery (SEB) with shared battery presents a complex landscape, requiring a detailed representation of the dynamic interactions between riders and e-bike batteries within the framework of a temporal graph. When users replace batteries, it becomes feasible for the same battery to appear at multiple exchange stations, thus enabling its reuse by a succession of different users across the system. In order to accurately capture the system's diverse characteristics, we utilize the notation $HG = {V, E}$, representing the heterogeneous graph input that encompasses the intricate relationships within the SEB ecosystem. Within this framework, the set of nodes is comprehensive, including both user nodes and battery nodes, which are defined as $V = V_{U} \cup V_B$, illustrating the system's dual aspects of interaction and utility. By interlinking users and batteries through edges, the structure of $HG$ naturally conforms to the characteristics of a bipartite graph, reflecting the two distinct but interconnected groups within the system. For clarity and further understanding, we present a simplified illustration of the SEB system in Fig.\ref{fig:SEBs}, which visualizes the conceptual framework discussed.


\subsection{Graph Neural Networks} \label{subsec: gnn}

Most Graph Neural Networks (GNNs) \cite{kipf2017semi} conform to the messing passing between neighboring nodes, which can be described as the following iteration functions:
\begin{equation}\label{equation: GNN-MP}
    m_{v}^{l+1} = f_{\theta}^{l} ( h_{v}^{l}, \{h_{u}^{l} | u \in \mathcal{N}_{v}\} )
\end{equation}
\begin{equation}
    h_{v}^{l+1} = \sigma^{l} ( g^{l}( h_{v}^{l},m_{v}^{l+1}  ) ), 
\end{equation}
where $f_{\theta}^{l}, \sigma^{l}, g^{l}$ denote parametric functions, i.e., neighborhood aggregation function, activation function (e.g. sigmoid, ReLU), and combination function (e.g. summation, mean), at the $l$-layer messaging passing on graphs. $\mathcal{N}_{v}$ denotes the neighborhood for node $v$ and $h_{v}^{l}$ denotes the hidden embedding for $v$. Such message passing  in Eq.\ref{equation: GNN-MP} will repeat $L$ times ($l \in \{1, 2, \ldots, L\}$) until converge. For the combinatorial optimization task in this paper, the information will only pass across the entire graph.

Taking GCN for example,  the message-passing function in  graph convolutional network \cite{kipf2017semi} is explicitly written as:
\begin{equation}\label{equation: GNN}
    h_{v}^{l+1} = \sigma \left( \mathbf{W}^l \sum_{u \in \mathcal{N}_{v}} \frac{h_{u}^{l}}{|\mathcal{N}_{v}|} + \mathbf{B}^l h_{v}^{l}  \right),
\end{equation}
where $\mathbf{W}^l$ and $\mathbf{B}^l$ are learnable parameters in $l$-layer.

\subsection{Transformer}

Transformer \cite{vaswani2017attention,nguyen2023mitigating} signifies a notable achievement in the realm of natural language processing (NLP) and machine learning. Departing from traditional recurrent or convolutional architectures, the Transformer leverages a self-attention mechanism to capture contextual dependencies across input sequences, enabling it to excel in tasks such as language translation, summarization, and question-answering. The model's innovative architecture eliminates sequential dependencies, allowing for parallelization during training and significantly accelerating processing times. 


For a given input vector $\mathbf{X} = [\mathbf{x}_1, \mathbf{x}_2, \ldots, \mathbf{x}_N ] \in \mathbb{R}^{N \times D}$, where $\mathbf{x}_i \in \mathbb{R}^{D}$, transform embeds $\mathbf{X}$ into the output $\mathbf{H}$ in the following two steps:

\noindent{\bf Step 1:}\;We project $\mathbf{X}$ into three matrices: $\mathbf{Q}$ (query matrix), $\mathbf{K}$ (key matrix), $\mathbf{V}$ (value matrix) via linear transformation as follows:
\begin{equation}
    \mathbf{Q} = \mathbf{X}\mathbf{W}_Q^{\mathrm{T}}, 
    \mathbf{K} = \mathbf{X}\mathbf{W}_K^{\mathrm{T}}, 
    \mathbf{V} = \mathbf{X}\mathbf{W}_V^{\mathrm{T}}
\end{equation}
\begin{equation}
\mathbf{Q} = [\mathbf{q}(1), \mathbf{q}(2), \ldots, \mathbf{q}(N)]
\end{equation}
\begin{equation}
    \mathbf{K} = [\mathbf{k}(1), \mathbf{k}(2), \ldots, \mathbf{k}(N)]
\end{equation}
\begin{equation}
    \mathbf{V} = [\mathbf{v}(1), \mathbf{v}(2), \ldots, \mathbf{v}(N)]
\end{equation}
where $\mathbf{W}_Q, \mathbf{W}_K \in \mathbb{R}^{D_{qk} \times D}$ and $\mathbf{W}_V \in \mathbb{R}^{D}$.

\noindent{\bf Step 2:}\;The output vector $\mathbf{H} = [\mathbf{h}(1), \mathbf{h}(2), \ldots, \mathbf{h}(N)] \in \mathbb{R}^{N \times D_{qk}}$ is then computed as follows:
\begin{equation}
    \mathbf{H} = {Softmax} \left(\frac{\mathbf{Q}\mathbf{K}^{\mathrm{T}}}{\sqrt{D_{qk}}}\right) \mathbf{V}.
\end{equation}
Our work refers to a Transformer that uses the classical softmax attention.

\section{Framework}\label{section:framework}

\subsection{Dynamic SEB scenario}\label{subsec:scenario}

The scenario of sharing e-bike battery is complex and we describe the interaction between users and batteries as a temporary graph at time $t$. When users replace the battery, the same battery may appear at different exchange stations and subsequently be utilized by different users. We use $HG^{(t)} = \{V^{(t)}, E^{(t)}\}$ to represent the input heterogeneous graph at time $t$. The set of nodes includes user nodes and battery nodes as $V^{(t)} = V^{(t)}_{U} \cup V^{(t)}_B$. Thus, graph neural networks (see Section \ref{subsec: gnn}) can readily embed $HG^{(t)}$. Edges are connected by users and batteries, and then $HG^{(t)}$ can be described as a bipartite graph.

\subsection{SEB-Transformer}

Here, we introduce the details of SEB-Transformer. Initially, the dataset is bifurcated into two classifications: the first comprises standard time series Euclidean data, while the second encompasses non-Euclidean data representing the temporal dynamic graph (see Section \ref{subsec:scenario}). Then we use SEB-Transformer to update the features on edges and nodes. We update the embeddings of edge features by GNNs as Eq.\ref{equation: GNN}. Also, we use a Transformer to update the original features to predict the range of e-bike battery. The above process can be formulated as follows:
\begin{equation}
    \mathbf{X}^{(t)}_1 = {GNN}(V^{(t)}_U, V^{(t)}_B, HG^{(t)})
\end{equation}
\begin{equation}
    \mathbf{X}^{(t)}_2 = {Transformer}(V^{(t)}_U, V^{(t)}_B, {\mathbf{X}}^{(t)}_0)
\end{equation}

\noindent Then, we can obtain the predicted labels as:

\begin{equation}
    \widehat{\mathbf{Y}}^{(t)}={MLP}\left(
    \hat{\mathbf{X}}^{(t)}
    \right)
\end{equation}
where $\hat{\mathbf{X}}^{(t)} = {\mathbf{X}}^{(t)}_0 \bigoplus \hat{\mathbf{X}}^{(t)}_1 \bigoplus \hat{\mathbf{X}}^{(t)}_2$.

\subsection{S${}^3$IM}

We give a genetic form for the strong structural similarity index on graph S${}^3$IM: $\mathbb{R}^n \times \mathbb{R}^n \to \mathbb{R}$ as follows:
\begin{equation}\label{equation: ssim}
    S{}^3IM(x, y) = f(r_1(x, y), r_2(x, y), r_3(x, y))
\end{equation}
\begin{equation}
f(r_1, r_2, r_3) = [r_1]^{\alpha} \cdot [r_2]^{\beta} \cdot [r_3]^{\gamma}
\end{equation}
where $\alpha, \beta, \gamma$ are parameters used to adjust the relative importance of the three terms. 

The S${}^3$IM term contains three components which will be presented in the next subsection with details. As the previous paper \cite{wang2004image} says, the similarity measure should obey the following conditions:
\begin{itemize}
    \item Symmetric: S${}^3$IM$(x, y)$ = S${}^3$IM$(y, x)$.
    \item Boundedness: S${}^3$IM$(x, y) \leq M$
    \item Unique maximization: S${}^3$IM$(x, y) = M$ if and only if $x=y$ in element wise ($x_i=y_i,i \in [n]$).
\end{itemize}

Next, we will introduce the details of each component $r_1, r_2, r_3$ shown in Eq.\ref{equation: ssim} and we call them (1) luminance term, (2) contrast term, and (3) structure term, respectively.

\noindent{\bf (1) luminance term $r_1$.}\;As the first term, we can define the luminance as follows
\begin{equation}
    r_1(x, y) = \frac{2 \mu_x \mu_y + C_1}{\mu_x^2 + \mu_y^2 + C_1}
\end{equation}
where $C_1 > 0$ is a constant to avoid the denominator being zero. Usually, we choose $C_1 = (K_1 L)$, $L$ is the length of the output and $K_1 \ll 1$ is a pretty small constant.

\noindent{\bf (2) contrast term $r_2$.}\;The second term, we can define the contrast function as 
\begin{equation}
    r_2(x, y) = \frac{2 \sigma_x \sigma_y + C_2}{\sigma_x^2 + \sigma_y^2 + C_2}
\end{equation}
where $C_2 = (K_2 L)^2$ and $K_2 \ll 1$.

\noindent{\bf (3) structure term $r_3$.}\;The third term, we can refer to the following form
\begin{equation}
    r_3(x ,y) = \frac{\sigma_{xy} + C_3}{\sigma_x \sigma_y + C_3}
\end{equation}
where 
\begin{equation}
    \sigma_{xy} = \frac{1}{n-1} \sum_{i=1}^n (x_i - \mu_x)(y_i - \mu_y)
\end{equation}

\noindent{\bf Overview.}\;The primary aim of this section is to furnish a comprehensive overview of the SEB-Transformer, delineating its core principles and operational framework. Prior to embarking on the prediction of battery range, we precisely define the dynamic scenario of Sharing E-Bike Battery (SEB) as a heterogeneous graph, which effectively captures the intricate web of interactions occurring between users and batteries. Following this, we proceed to develop a structural Transformer model that meticulously incorporates topological information, achieving a seamless integration of both graph-based and sequential data, thereby enriching the model's predictive capabilities. Furthermore, to augment the model's ability to recognize global structural information, we implement the advanced regularization technique S${}^3$IM, which significantly improves the precision of our predictions. Throughout the training phase, our objective shifts towards minimizing the discrepancy between the actual battery range $\mathbf{Y}^{(t)}$ and its predicted counterpart $\widehat{\mathbf{Y}}^{(t)}$, over the entirety of the temporal spectrum under consideration. The optimization strategy employed for our model is delineated as follows, encompassing a detailed methodology aimed at refining the accuracy of our predictive framework:
\begin{equation}
    \min \sum_{t=1}^T \left( \frac{
    \|\widehat{\mathbf{Y}}^{(t)} - \mathbf{Y}^{(t)}\|^2
    }{|\mathbf{Y}^{(t)|}} + 
    {S{}^3IM}(\widehat{\mathbf{Y}}^{(t)}, \mathbf{Y}^{(t)}) \right).
\end{equation}

\section{Experiments}\label{section:exp}

This section provides a detailed exposition of our experiments. Section \ref{subsec: dataset} presents a real-world dataset collected by our team, accompanied by an in-depth analysis of its statistical information from various perspectives. Section \ref{subsec: baseline} outlines several competitive baselines that serve as benchmarks for comparison, aiming to surpass their performance. Section \ref{subsec: results} highlights the main results assessed by our model.

\subsection{Dataset}\label{subsec: dataset}

\begin{figure}
    \centering
    \includegraphics[width=.9\linewidth]{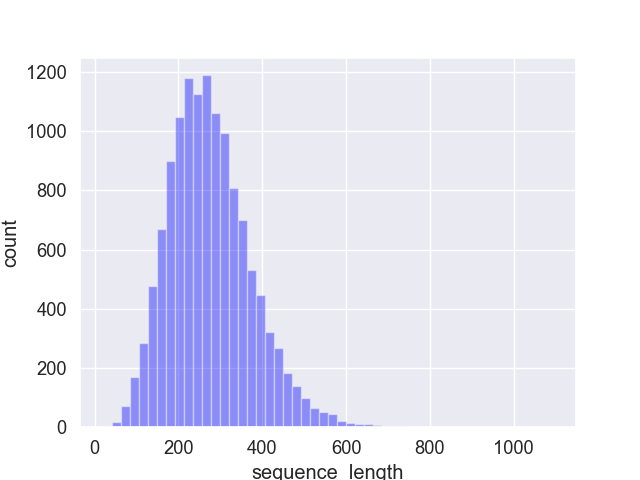}
    \caption{\small{Statistical visualization.}} \label{fig:Statistical} 
    \vspace{-10pt}
\end{figure}

For our experimental investigations, we utilize a real-world dataset that has been meticulously collected by our team. Our dedicated engineering team meticulously curated a dataset comprising 16,000 order samples, each enriched with 64 distinct timing data points. 
Subsequently, we provide an overview of statistical information pertaining to our dataset. In Fig.\ref{fig:Statistical}, we provided visualizations depicting sequence length information. Observation of the dataset reveals that it approximates a normal distribution, with an average value of 275, thereby suggesting that the data we gathered is both reasonable and reflective of societal dynamics. 




\begin{figure*}
    \centering
    \includegraphics[width=1.0\linewidth]{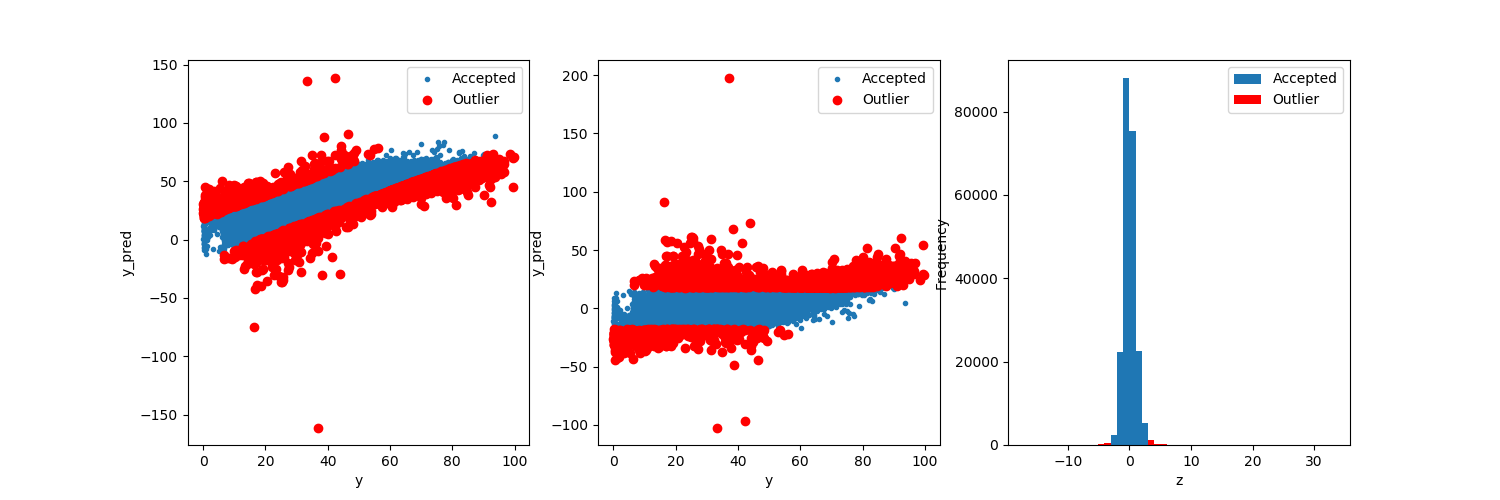}
    \caption{\small{Analysis of the outliers in our data.}} \label{fig:outlier} 
    \vspace{-10pt}
\end{figure*}

\subsection{Baseline}\label{subsec: baseline}

Within this section, we present an overview of the competitive baseline methods utilized in our study, namely SVR, LR, XGBoost, MLP, and the vanilla Transformer model. These methods are implemented for our task.

\begin{itemize}
    \item {\bf Support Vector Regression (SVR)}\;\cite{cervantes2020comprehensive,zhang2012support} stands out as a robust machine learning method for regression challenges, especially adept at managing datasets where the variables exhibit intricate interconnections. 
    \item {\bf Linear Regression (LR)}\;\cite{ahmed2022electric} is a fundamental statistical method utilized in machine learning and data analysis for establishing a connection between a dependent variable and one or more independent variables. LR presupposes a direct linear connection between the input factors and the outcome, depicted as a straight line within a multidimensional space. The model calculates the coefficients for the linear equation that most closely aligns with the observed data points, by reducing the sum of the squared differences (residuals).
    \item {\bf eXtreme Gradient Boosting}\;\cite{chen2016xgboost} is an effective and robust machine learning technique renowned for its effectiveness in predictive modeling and classification tasks. It belongs to the ensemble learning family, specifically boosting algorithms, which sequentially combine weak learners to create a robust and accurate model. XGBoost utilizes a gradient boosting framework, employing decision trees as base learners. It excels in handling structured data and is known for its speed, scalability, and accuracy. XGBoost incorporates regularization techniques to prevent overfitting and provides features for fine-tuning model performance, such as cross-validation and early stopping. 
    \item {\bf Multilayer Perceptron (MLP)}\cite{hasib2021driving} represents a basic artificial neural network structure, featuring several layers of linked neurons, including an input layer, at least one hidden layer, and an output layer.
    \item {\bf Transformer}\;\cite{vaswani2017attention} is a groundbreaking architecture in deep learning that has notably propelled advancements in natural language processing (NLP) and related sequence-dependent tasks. Unlike previous recurrent or convolutional models, the Transformer relies entirely on self-attention mechanisms to capture dependencies between input and output tokens in parallel. 
\end{itemize}



\subsection{Main Results}\label{subsec: results}

\begin{figure}[t]
    \centering
    \includegraphics[width=\linewidth]{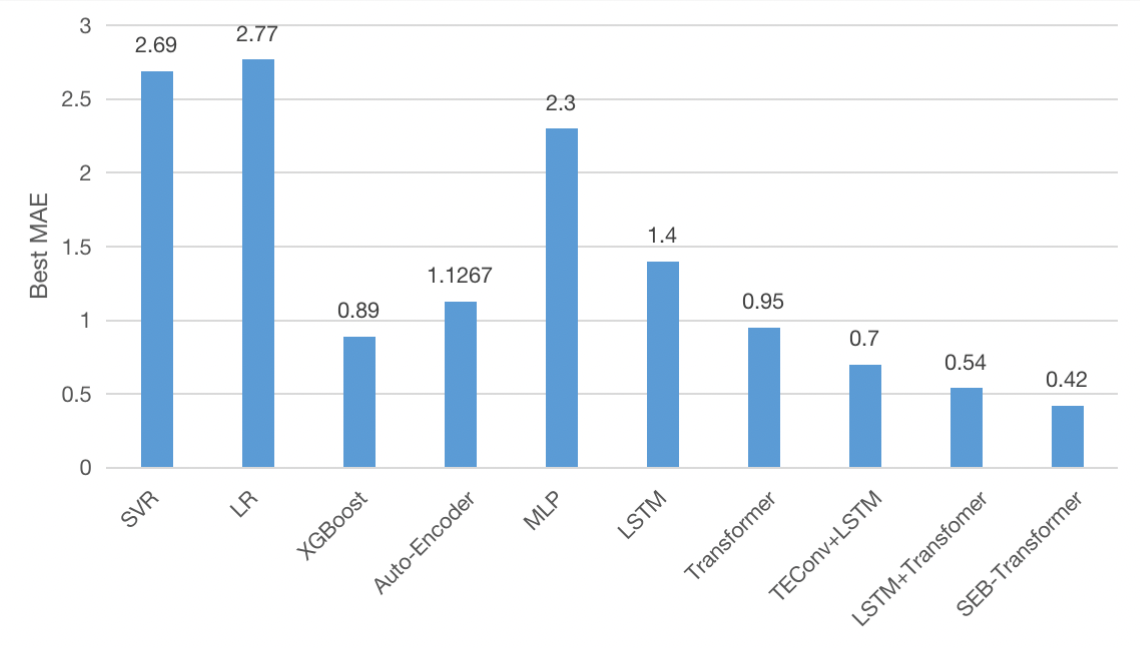}
    \caption{\small{Illustration of MAE for the best performance.}} \label{fig:best} 
    \vspace{-10pt}
\end{figure}
\begin{figure}[t]
    \centering
    \includegraphics[width=\linewidth]{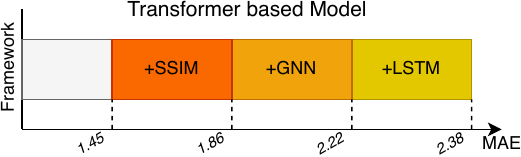}
    \caption{\small{Demonstration of framework improvements.}} \label{fig:improvements} 
    \vspace{-10pt}
\end{figure}

The experimental results have been succinctly summarized in Table \ref{tab:mae}. The last two models represent variants of the SEB-Transformer architecture. An additional regularization term, denoted as S${}^3$IM, has been incorporated into the SEB-Transformer${}^*$ architecture. The aforementioned models serve as ten competitive baselines, providing a robust benchmark for evaluation. Remarkably, our model surpasses these baselines by a substantial margin. Furthermore, Fig.\ref{fig:improvements} has been included to visually illustrate the effects of each improvement. This visual representation effectively demonstrates the impact of individual improvements. Additionally, the Fig.\ref{fig:best} showcases the best MAE achieved by our model. This presentation serves to underscore that our model consistently outperforms others even in dynamic stream conditions.

\begin{figure*}[h]
\mbox{
        \includegraphics[width=0.33\linewidth]{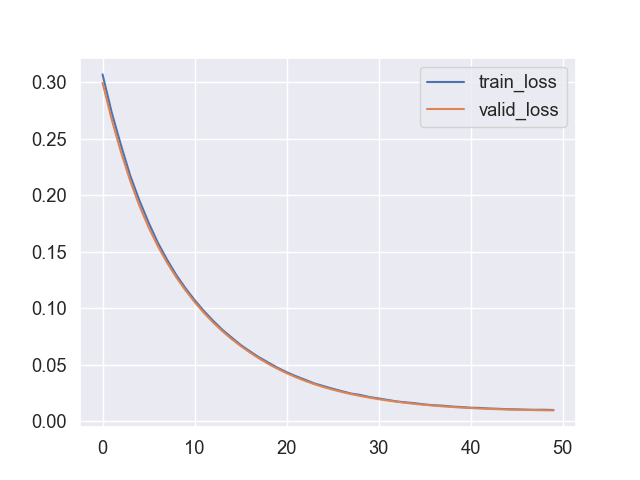}
        \includegraphics[width=0.33\linewidth]{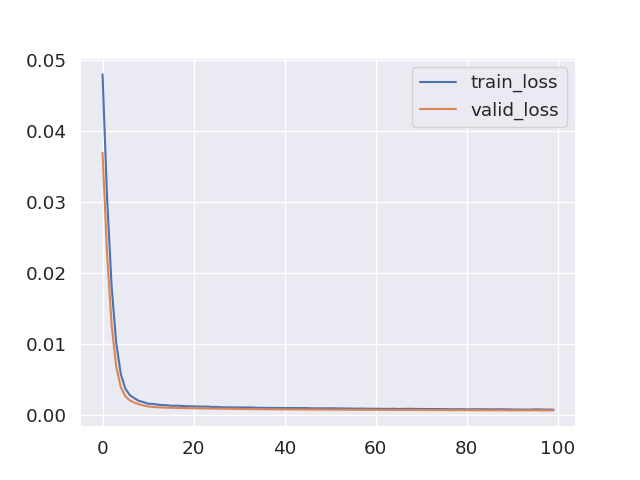}
        \includegraphics[width=0.33\linewidth]{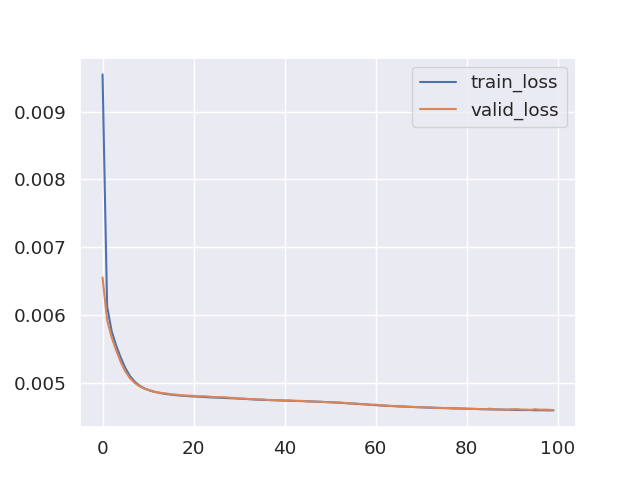}
    }
        
    \mbox{
    \includegraphics[width=0.33\linewidth]{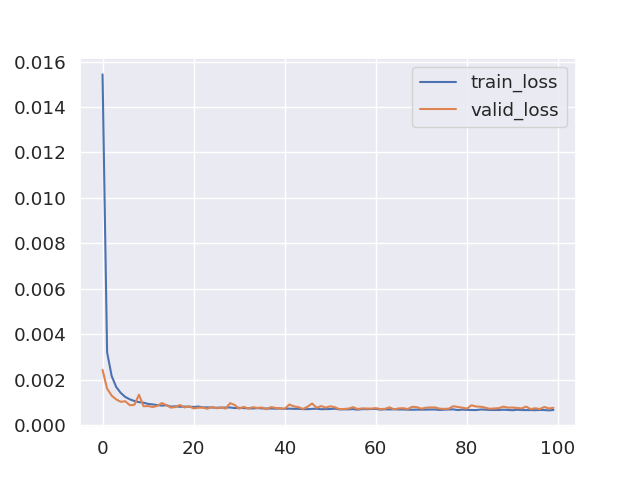}
        \includegraphics[width=0.33\linewidth]{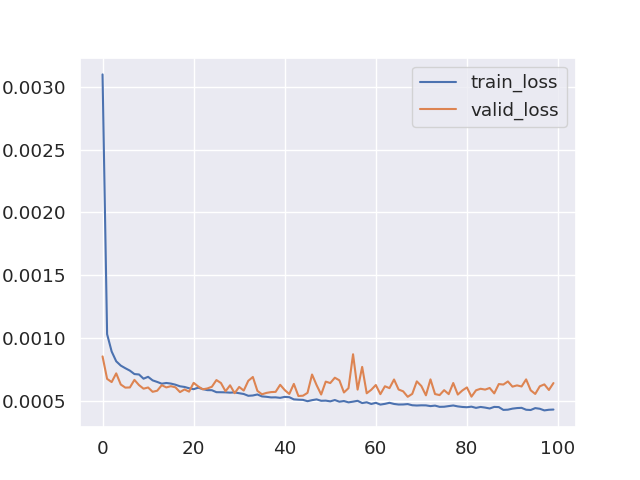}
        \includegraphics[width=0.33\linewidth]{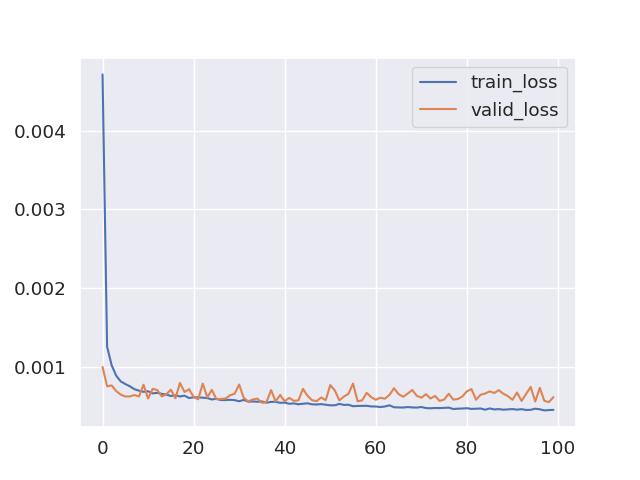}
    }
        \caption{\small{Training loss among various models: (top line) auto-encoder, lstm, mlp, and (bottom line) Transformer, Transformer plus lstm, SEB-Transformer.}} \label{fig:training-loss} 
        \vspace{-10pt}
    \end{figure*}

Fig.\ref{fig:training-loss} presents a visual depiction of the loss curve as observed throughout the training and validation phases, providing a graphical representation of the model's performance over these periods. When compared to the baseline models, our framework demonstrates a decrease in loss, indicating its superior performance in reducing the discrepancy between predicted outcomes and actual values. 
This improvement underscores the framework's superior efficacy in aligning predictions closely with real-world data.


\begin{table}[t]
\small
\caption{\small{MAE on the battery range prediction task}}
    \label{tab:mae}
    \centering
    \begin{tabular}{|l|l|}
    \hline {Model } & {MAE } $\downarrow$ \\
\hline {SVR } & 4.7564 $\pm$ 2.0711 \\
{LR} & 4.7645 $\pm$ 1.9910 \\
{XGBoost} & 5.1529 $\pm$ 4.2586 \\
{Auto-Encoder} & 5.3849 $\pm$ 4.2582 \\
{MLP} & 4.3606 $\pm$ 2.0636 \\
{LSTM} & 3.1380 $\pm$ 1.7353 \\
{Transfomer} & 2.3772 $\pm$ 1.4306 \\
{TEConv+LSTM} & 2.2240 $\pm$ 1.5272 \\
{TEConv+Transformer} & 1.9908 $\pm$ 1.7498 \\
{LSTM+Transfomer} & 1.8642 $\pm$ 1.3228 \\
\hline
{SEB-Transformer} & {1.5020} $\pm$ 1.2281 \\
{SEB-Transformer}${}^*$ & \textbf{1.4552} $\pm$ 1.0332
\\
\hline
    \end{tabular}
    \vspace{-10pt}
\end{table}

\section{Application with Web Service}\label{section: application}


Within Sharing E-Bike Battery (SEB), the app and web service are key to user experience and system efficiency. Future improvements promise to enhance SEB functionality and accessibility, fostering wider use and greater satisfaction. By addressing these application perspectives, SEB have the opportunity to enhance user satisfaction, optimize system efficiency, and contribute to the widespread adoption of sustainable transportation solutions.

\begin{itemize}
    \item {\bf User Interface Design.} Prioritizing user-centric design principles \cite{stone2005user} in mobile applications is essential for offering intuitive interfaces that enable users to access battery status information, locate nearby SEB stations, and plan routes based on predicted battery ranges. Adding features like interactive maps, customizable ride options, and live updates could greatly improve the user experience and encourage more frequent use of SEB services.
    These enhancements can contribute to a more engaging and satisfying experience for users, thereby promoting greater adoption and usage of SEB services.
    \item {\bf Predictive Range Estimation.} The integration of battery range prediction algorithms \cite{enthaler2015method} directly into the application interface aims to empower users by providing them with accurate estimations of remaining battery life for their planned journeys. This integration could involve the display of estimated range, considering factors such as user weight, terrain, and historical usage patterns. 
    \item {\bf Ride Optimization Features.} The implementation of ride optimization features within the application aims to improve user convenience and system efficiency. These features may encompass smart routing algorithms capable of suggesting optimal paths based on available battery range and user preferences. Additionally, real-time alerts and notifications will be integrated to remind users to recharge or swap batteries as needed, further enhancing the overall user experience and system efficiency.
    \item {\bf Data Visualization and Analytics.} Utilizing data visualization techniques \cite{andrienko2020big} within the application interface to present battery usage statistics, ride history, and system performance metrics in a clear and comprehensible manner is essential. By providing users with insights into their usage patterns, environmental impact, and potential cost savings, we can foster greater awareness and engagement with SEB services.
    \item {\bf Integration with Urban Mobility Ecosystem.} Facilitating seamless integration with existing urban mobility ecosystems involves ensuring interoperability with public transit systems, ride-sharing platforms, and navigation services. This may entail the provision of multi-modal trip planning capabilities, integration of payment systems for seamless fare integration, and enabling cross-platform connectivity to enhance overall transportation accessibility and convenience.
\end{itemize}

\section{Discussion and Conclusion}\label{section:conclusion}

\noindent{\bf Why We Need Sharing E-Bike Battery.}\; (1) Bridging Expectation Gaps: A notable discrepancy exists between user expectations and the actual remaining battery range of SEBs, especially in urgent needs for accessible SEBs. (2) Operational Efficiency and Energy Conservation: Accurate battery range prediction and strategic placement of exchange stations can significantly enhance route planning, operational efficiency, and energy conservation. (3) Promoting Sustainable Transportation: By overcoming limitations associated with traditional charging infrastructure, sharing e-bike batteries can lead to a more efficient and environmentally friendly transportation ecosystem.

\noindent{\bf Conclusion.}\;In this paper, we have conceptualized the scenario involving Sharing E-Bike Battery (SEB) as a dynamic heterogeneous graph. Concurrently, we introduce the innovative SEB-Transformer, which is specifically designed for the purpose of battery range prediction. Furthermore, it is noteworthy that our model significantly outperforms the conventional Transformer models, indicating a substantial improvement in predictive performance. The contributions of our work represent a notable advancement within the domain of web services, characterized by a dual achievement: the reduction of carbon emissions and the enhancement of user satisfaction, thereby marking a pivotal step forward in sustainable urban mobility. We propose a novel, task-oriented model termed as SEB-Transformer, which employs a structural Transformer-based methodology specifically devised for the accurate prediction of battery range in SEBs. Utilizing the prediction, we can dynamically modify optimal cycling routes for users in real-time, taking into account the locations of charging stations. Experimental results shows that our model outperforms nine other competitive baseline models on real-world datasets, with a significant reduction in the MAE (Mean Absolute Error) metric compared to the basic Transformer model.

\section*{Acknowledgment}

This work was partially supported by the National Key R\&D Program of China (2023YFB4502400), the Strategic Priority Research Program of the Chinese Academy of Sciences (XDB0680101), the NSFC (62376064, U2336202), and the CAS Project for Young Scientists in Basic Research (YSBR-008).

\bibliographystyle{IEEEtran}
\bibliography{ref}

\end{document}